\documentclass[conference]{IEEEtran}
\IEEEoverridecommandlockouts
\usepackage{cite}
\usepackage{amsmath,amssymb,amsfonts}
\usepackage{algorithmic}
\usepackage{graphicx}
\usepackage{textcomp}
\usepackage{xcolor}
\usepackage{threeparttable}
\def\BibTeX{{\rm B\kern-.05em{\sc i\kern-.025em b}\kern-.08em
    T\kern-.1667em\lower.7ex\hbox{E}\kern-.125emX}}
\begin{document}

\makeatletter
\newcommand{\linebreakand}{%
  \end{@IEEEauthorhalign}
  \hfill\mbox{}\par
  \mbox{}\hfill\begin{@IEEEauthorhalign}
}
\makeatother

\title{Deep Learning-Based Electricity Price Forecast for Virtual Bidding in Wholesale Electricity Market}

\author{

\IEEEauthorblockN{Xuesong Wang, Sharaf K. Magableh, Oraib Dawaghreh, Caisheng Wang*\thanks{*Corresponding author.}}
\IEEEauthorblockA{\textit{Department of Electrical and Computer Engineering}\\
\textit{Wayne State University}\\
Detroit, MI, USA\\
\{xswang, sharaf.magableh, oraib.dawaghreh, cwang\}@wayne.edu
}

\linebreakand

\IEEEauthorblockN{Jiaxuan Gong}
\IEEEauthorblockA{\textit{Northville High School}\\
Northville, MI, USA\\
jacksongong2017@gmail.com}

\and

\IEEEauthorblockN{Zhongyang Zhao}
\IEEEauthorblockA{\textit{Greenwich Commodities LLC}\\
Troy, MI, USA\\
zzhao@greenwichcm.com
}
\and
\IEEEauthorblockN{Michael H. Liao}
\IEEEauthorblockA{\textit{Greenwich Commodities LLC}\\
Troy, MI, USA\\
mliao@greenwichcm.com
}

}

\maketitle

\begin{abstract}
Virtual bidding plays an important role in two-settlement electric power markets, as it can reduce discrepancies between day-ahead and real-time markets. Renewable energy penetration increases volatility in electricity prices, making accurate forecasting critical for virtual bidders, reducing uncertainty and maximizing profits. This study presents a Transformer-based deep learning model to forecast the price spread between real-time and day-ahead electricity prices in the ERCOT (Electric Reliability Council of Texas) market. The proposed model leverages various time-series features, including load forecasts, solar and wind generation forecasts, and temporal attributes. The model is trained under realistic constraints and validated using a walk-forward approach by updating the model every week. Based on the price spread prediction results, several trading strategies are proposed and the most effective strategy for maximizing cumulative profit under realistic market conditions is identified through backtesting. The results show that the strategy of trading only at the peak hour with a precision score of over 50\% produces nearly consistent profit over the test period. The proposed method underscores the importance of an accurate electricity price forecasting model and introduces a new method of evaluating the price forecast model from a virtual bidder's perspective, providing valuable insights for future research.
\end{abstract}


\section{Introduction}

The increasing penetration of renewable energy sources in power grids has amplified the need for accurate electricity price forecasting to enhance market efficiency and mitigate uncertainty for participants. In the wholesale electricity market, virtual bidding is an essential mechanism that allows market participants to profit from discrepancies between day-ahead (DAM) and real-time market prices, thereby helping to align these two markets more closely. This practice requires an accurate forecasting model to predict price spreads, which is essential for minimizing risk and maximizing gains. In recent years, deep learning has emerged as a powerful tool in energy forecasting due to its ability to model complex temporal dependencies and capture intricate patterns within large datasets.

Significant progress has been made in forecasting day-ahead and real-time electricity prices using advanced machine learning techniques. Deep learning models, such as Long Short-Term Memory (LSTM) and Convolutional Neural Network (CNN), developed in \cite{R1}, have shown to outperform traditional statistical methods, e.g., Autoregressive Integrated Moving Average (ARIMA) \cite{R13}, in terms of accuracy for predicting the day-ahead price of the ERCOT (Electric Reliability Council of Texas) South Load Zone. Similarly, LSTM and Temporal Convolutional Networks (TCN) were proposed to predict day-ahead prices on the Hungarian HUPX market \cite{R12}. Focusing on the day-ahead prices in the German-Luxembourg bidding zone, the LSTM models developed in \cite{R6} produced accurate forecasts of electricity prices and price volatility, where load and renewable forecasts were utilized.
Other machine learning models have also been developed for real-time price forecasting. For instance, Random Forest and LSTM models were used to predict the real-time prices of the New South Wales electricity market in Australia and the Singapore market \cite{R4}. These studies demonstrate the great potential of applying machine learning techniques to day-ahead and real-time price forecasting.

The profit from virtual bidding relies on the price difference between day-ahead and real-time markets. Recent studies have explored different approaches to optimize bidding decisions based on price forecasts. For example, four machine learning models were proposed in \cite{R2} to predict spread spikes for the Long Island zone of the New York Independent System Operator (NYISO) and two heuristic trading strategies were proposed to validate the model performance. Similarly, a mixture density network was proposed in \cite{R3} to predict the price spread in the ISO-NE (New England) market, which was utilized to develop a machine learning framework for algorithmic trading where a budget- and risk-constrained portfolio optimization problem was solved. In \cite{R5}, LSTM and Sequence-to-Sequence (Seq2Seq) models were proposed to predict the price spread for the PJM (Pennsylvania-New Jersey-Maryland Interconnection) market. However, the model proposed in \cite{R5} only forecasted price differences for the next hour, which is not aligned with the two-settlement market practice in the US. As part of the analysis of algorithmic trading, an LSTM model was developed to predict the price spread for PJM, CAISO (California Independent System Operator), and ISO-NE \cite{R7,R10}. ERCOT, the only major energy-only market in the United States, experiences high price volatility due to its reliance on real-time price signals. This volatility, as shown in Fig.~\ref{price_spread}, is amplified by increasing renewable energy penetration. Despite its importance, virtual bidding in ERCOT remains underexplored, making accurate forecasting crucial for effective strategies.

\begin{figure}[b]
    \centering
    \includegraphics[width=1\linewidth]{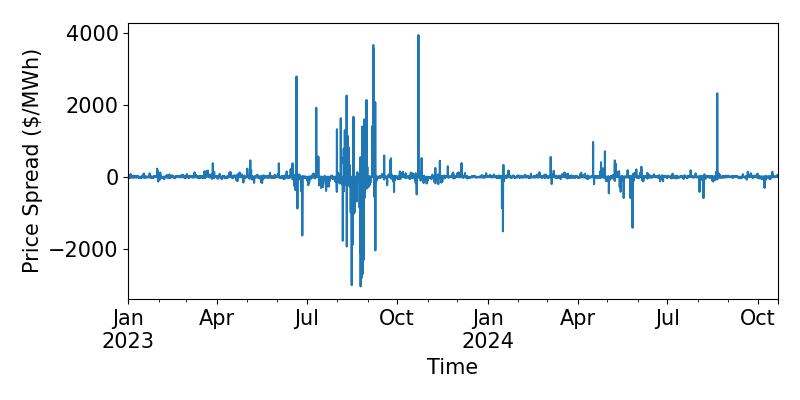}
    \caption{Price spread between real-time price and day-ahead price of system lambda in ERCOT market.}
    \label{price_spread}
\end{figure}

This paper presents a novel approach for forecasting the price spread between the real-time Security Constrained Economic Dispatch (SCED) and the DAM system lambda prices in the ERCOT market. Specifically, a Transformer-based deep learning model is proposed, leveraging various features, including time information, load forecasts, solar generation forecasts, and wind generation forecasts. The model aims to classify the price spreads into different ranges, enabling effective decision-making in virtual bidding scenarios. It is trained using the information available by the time of bid submission and updated every week, reflecting realistic application scenarios. Based on the prediction results, various trading strategies are developed with a specific focus on maximizing profitability under realistic market conditions. Through backtesting on ten months of data in 2024, this work demonstrates the efficacy of the proposed price spread forecast model and the virtual bidding strategies.

The key contributions of this work are as follows:

\begin{itemize}
    \item A Transformer-based deep learning model is proposed to forecast the spread between SCED and DAM prices in the highly volatile ERCOT market.
    \item Different training settings are compared in terms of traditional evaluation metrics and the cumulative net profit.
    \item Multiple trading strategies are developed based on the model's prediction and assessed in realistic market scenarios.
\end{itemize}

The remainder of the paper is organized as follows: Section II discusses the methodology and model design. Section III presents the results and analysis, including trading performance evaluation. Section IV concludes the paper with a discussion of limitations and future work.

\section{Methodology}

\subsection{Model Design}

The forecasting model is built using a Transformer-based deep learning architecture, which is particularly suited for time-series prediction tasks due to its attention mechanism that captures complex temporal dependencies. The main components of the model include a linear encoder, a positional encoding layer, several transformer encoder layers, and a decoder layer, as shown in Fig.~\ref{model_arch}. The linear encoder projects the inputs into a high dimensional space; the positional encoding layer adds positional information to the input feature of the transformer encoder layers; the decoder layer maps the vectors from the high dimensional space into the output space. The time step of the model is one day. Today is referred to as day T, tomorrow as T+1 which is the prediction target. The input features used in the model for each day are as follows:

\begin{figure}[b]
    \centering
    \includegraphics[width=0.6\linewidth]{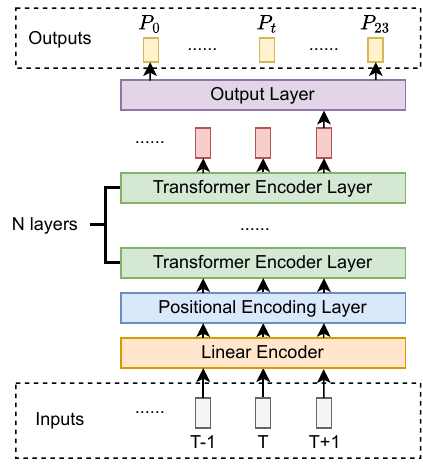}
    \caption{Model architecture, where the output $P_t$ represent the probability distribution of the price spread of hour 0:00 (In ERCOT, it needs to be converted into Hour Ending 1:00).}
    \label{model_arch}
\end{figure}

\begin{itemize}
    \item Time information: holiday information, year, month, day of month, and day of week.
    \item Load forecast: the load forecast for each hour of the day provided by ERCOT for different weather zones.
    \item Solar generation: the solar generation forecast for each hour of the day provided by ERCOT for different geographical zones.
    \item Wind generation: the wind generation forecast for each hour of the day provided by ERCOT for different geographical zones.
    \item Price spread: the difference between SCED and DAM system lambda prices.
\end{itemize}

Renewable energy has been increasing in ERCOT market in recent years. For example, the solar generation in recent years is shown in Fig.~\ref{solar_cop}. Therefore, including time information in the model inputs provides trend information implicitly. The price spread information for day T is not available by the time of bid submission thus is replaced by a zero vector for day T.

\begin{figure}[b]
    \centering
    \includegraphics[width=1\linewidth]{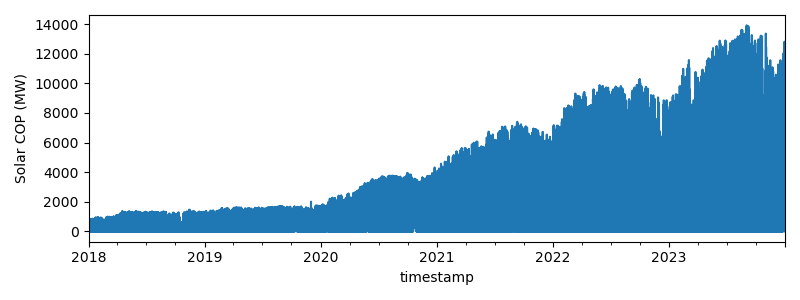}
    \caption{Annual solar generation Current Operation Plan (COP) trends from 2018 to 2024.}
    \label{solar_cop}
\end{figure}

The price spread is quantized into 5 ranges: $(-\inf, -12)$, $[-12,-5)$, $[-5,5)$, $[5,12)$, $[12,\inf)$. The threshold of \$5.00 is chosen as the break even point of the virtual bidding considering the uplist cost. The threshold of \$12.00 is about the median of the price spreads that is outside \$5.00. Therefore, the outputs of the model are the probability distribution over these five categories for 24 hours of day T+1.

\subsection{Data Preprocessing and Dataset Split}

The load forecast and renewable generation forecast are scaled using their respective interquartile ranges (IQR) to enhance model convergence and stability during training. The IQR-based scaling ensures that the features are on a comparable scale, which aids in gradient-based optimization and facilitates faster convergence. Additionally, IQR scaling is robust to outliers, which is particularly beneficial given the high variability of electricity load and renewable generation data. By focusing on the central portion of the data, the model can better capture typical patterns, thereby improving its generalizability to real-world scenarios. The month, day of the month, and day of the week are represented using sine and cosine transformations to effectively capture cyclical patterns in the data. The data set spans from Jan. 1, 2023 to Oct. 20, 2024. The data set in 2024 is used as test set. The model is validated using walk-forward validation to reflect realistic use case, i.e., the model is updated every 7 days, allowing the model to learn from the most recent data and adapt to changing market conditions. The initial train set covers from Jan. 1 to Dec. 23, 2023, validation set from Dec. 24 to Dec. 30, 2023, and test set from Jan. 1 to Jan. 7, 2024. There is a one-day gap between the validation set and test set because in real-world scenarios in order to train the model using the data up to Dec. 30, 2023, the model has to be trained on Dec. 31 or later, in such case the model can only be used for bid submission for Jan. 1 of 2024 or later.

\subsection{Model Training}

This study aims to investigate the impact of three training settings on the model's performance:

\begin{itemize}
    \item Lagging: how many days of historical information to include as inputs. For instance, with lagging=2, the inputs include information of day T-1, T, and T+1. In this work, laggings of 1, 2, and 3 are compared.
    \item Train set size: how many days of historical data are used as training data for each week. In this study, four settings are compared, i.e., 90, 180, 360 days of data and all historical data.
    \item Finetuning or training from scratch each week. In each week, the model can either be finetuned or trained from scratch.
\end{itemize}

The models are trained with cross-entropy loss. For each setting, four hyper parameters are searched using Ray Tune \cite{R11} via 50 trials: number of transformer encoder layer, the dropout rate of the positional encoding layer, the learning rate and the weight decay. For each week, the model with the lowest validation loss is used to predict the results on test set. The metrics used include accuracy, precision, recall, and F1 score.

\subsection{Trading Strategy Design Principle}

Besides the typical metrics of classification performance, the models are also evaluated by cumulative net profit in virtual bidding settings. The profitability of each strategy is evaluated through backtesting using historical data. The results of these strategies are compared to establish the most effective approach under realistic market conditions. A few assumptions are made when designing and evaluating the trading strategies:
\begin{itemize}
    \item The uplift cost of trading 1 MWh energy is \$5.00.
    \item The bids and offers always get cleared by placing the price at price cap or price floor, details please refer to \cite{R7}.
    \item There is a budget limit each day, i.e., only 1 MWh is traded per day, but it may be split in multiple hours evenly if multiple hours are selected to trade in a single day.
    \item The bids and offers will not change the system lambda.
\end{itemize}

\section{Results and Analysis}

In this section, we present the results of the forecasting model and evaluate the performance of the proposed trading strategies.

\subsection{Model Performance Evaluation}

\begin{table*}[bthp]
    \caption{Performance Metrics of Models under Different Settings}
    \centering
    \begin{threeparttable}
    \begin{tabular}{c|c|c|c|c|c|c|c|c|c|c|c|c|c|c}\hline
         Model & L & S & F & Acc & Pre & Rec & F1 & T1 & T2 & T3 & T4 & T5 & T6 & T7 \\
         \hline
         A & 1 & 90  & No  & 0.499 & 0.345 & 0.285 & 0.278 & \textbf{1246} & 777 & 1819 & 1270 & 238 & 5134 & 14957 \\
         B & 2 & 90  & No  & 0.499 & 0.333 & 0.284 & 0.273 & 1001 & -1213 & 1290 & 718 & -262 & 5134 & 14957 \\
         C & 3 & 90  & No  & 0.496 & \textbf{0.349} & 0.286 & \textbf{0.282} & 447 & -177 & 142 & 671 & -633 & 5134 & 14957 \\
         D & 1 & 180 & No  & 0.496 & 0.326 & 0.286 & 0.275 & 145 & 525 & -221 & 89 & -680 & 5134 & 14957 \\
         E & 1 & 360 & No  & 0.501 & 0.322 & 0.293 & 0.279 & -723 & 722 & -115 & 882 & -8 & 5134 & 14957 \\
         F & 1 & All & No  & \textbf{0.508} & 0.335 & \textbf{0.296} & 0.275 & 137 & \textbf{1070} & 1648 & \textbf{2258} & \textbf{1369} & 5134 & 14957 \\
         G & 1 & 90  & Yes & 0.503 & 0.340 & 0.279 & 0.270 & -235 & 453 & \underline{\textbf{2482}} & 1577 & 569 & 5134 & 14957 \\
         \hline
    \end{tabular}
    \begin{tablenotes}
            \item[1] L: lagging, S: train set size, F: finetuning, Acc: accuracy, Pre: precision, Rec: recall
            \item[2] T1 - T7: cumulative net profit of trading strategy T1 to T7. 
        \end{tablenotes}
    \label{perf_metrics}
    \end{threeparttable}
\end{table*}

The performance metrics under different training settings are shown in Table~\ref{perf_metrics}, where the precision, recall and F1 are macro scores. Overall, the performance differences between training settings are minimal, with model \textbf{C} showing slight advantages in precision and F1 score, while model \textbf{F} excels in accuracy and recall. The houly confusion matrix for model \textbf{G} is shown in Fig.~\ref{confusion_matrix}, which is very similar to that of the other models. The results showed that for most hours the model predicted the correct class, which is $[-5,5)$, indicating there is no obvious trading opportunity, except hour 19, where the precision for the class $(-\inf, -12)$ reaches or exceeds 50\%. In the context of virtual bidding, precision matters more than recall. Precision measures the reliability of the model, whereas recall indicates the capability of capturing potential trading opportunity. If the bidders trust a model with low precision, they would incur a large loss. On the contrary, bidders would miss the opportunities of making profit by trusting a model with low recall. Based on the analysis above, hour 19 is selected to build most of the trading strategies to maximize profit while balancing potential loss.

\begin{figure*}
    \centering
    \includegraphics[width=0.94\linewidth]{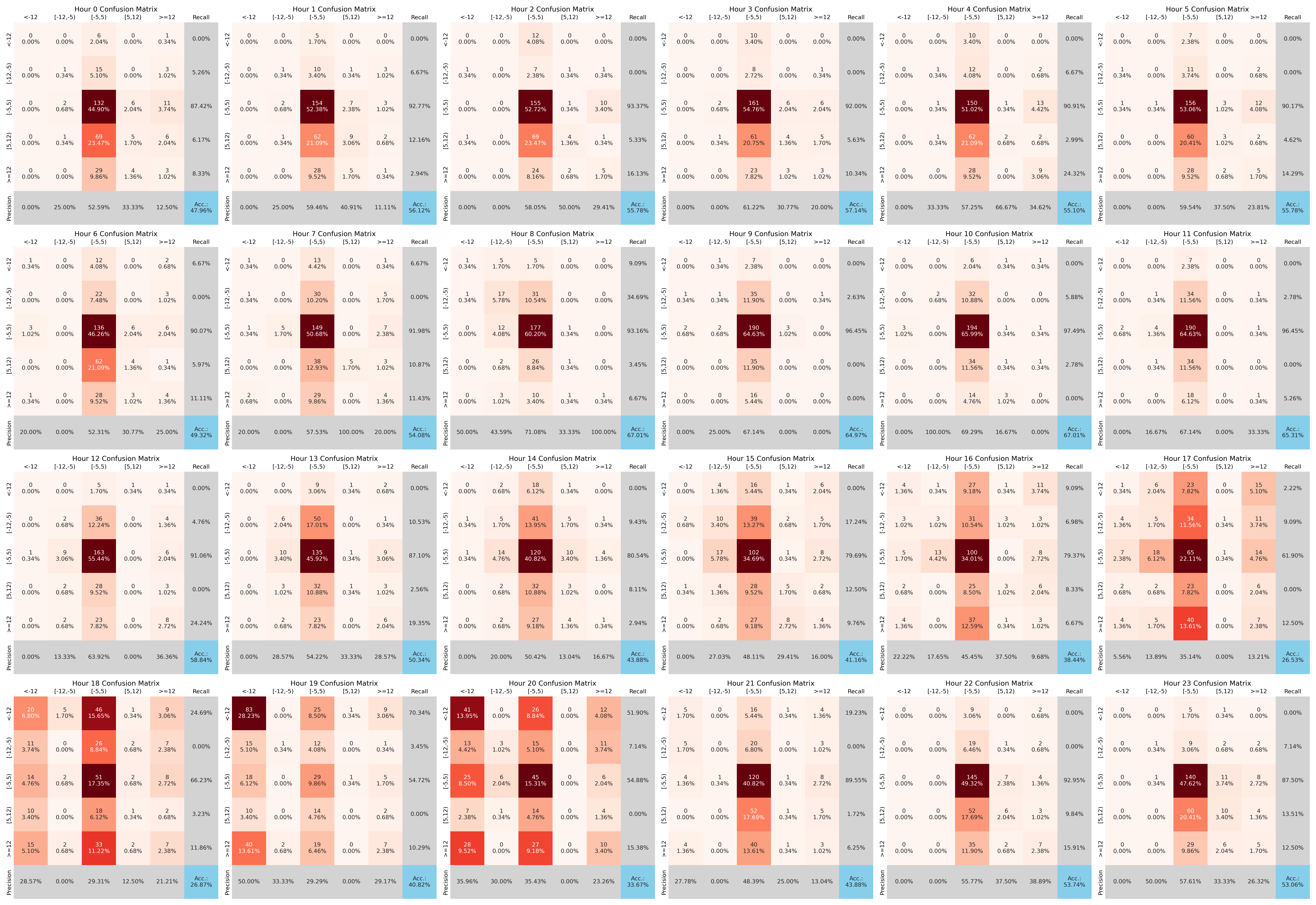}
    \caption{Confusion matrix of the best model for each hour. Precision, recall, and accuracy are best viewed when zoomed in.}
    \label{confusion_matrix}
\end{figure*}

\subsection{Trading Strategy Evaluation}

This work proposed seven trading strategies as follows:

\begin{itemize}
    \item \textbf{T1}: trades are executed for all hours based on the model's prediction.
    \item \textbf{T2}: trades are executed for all hours based on the model's prediction but only when the predicted probability is greater than 50\%.
    \item \textbf{T3}: trades are executed for hour 19 only based on model's prediction.
    \item \textbf{T4}: trades are executed for hour 19 only based on model's prediction but only when the predicted class is $(-\inf,-12)$.
    \item \textbf{T5}: trades are executed for hour 19 only based on model's prediction but only when the predicted class is $(-\inf,-12)$ and the predicted probability is greater than 50\%.
    \item \textbf{T6}: trades are executed for all hours based on the ground truth. This is an ideal case which is never going to happen in real world.
    \item \textbf{T7}: trades are executed for hour 19 only based on the ground truth. Like T6, this is an ideal case that is impractical in real-world scenarios.
\end{itemize}

The cumulative net profits over the test period for all the models under all trading strategies are listed in Table~\ref{perf_metrics}. As can be seen from Table~\ref{perf_metrics}, even though model \textbf{C} showed advantage on precision and F1 score, it didn't perform well under any of the strategies. While model \textbf{F} performed well under three different strategies, aligned with its advantages on accuracy and recall, the most profitable strategy is from model \textbf{G}. The mismatch between the trading profit and the traditional metrics of classification models highlights the importance of measuring the model's performance in the end-to-end application.

Fig.~\ref{cumulative_profit} shows the cumulative profit curves for each strategy based on model \textbf{G}, illustrating the performance over the test period, i.e., Jan. 1 to Oct. 20 of 2024. Several observations can be made from the figure:
\begin{itemize}
    \item Strategy \textbf{T7} has higher profit than \textbf{T6}, which can be attributed to budget constraints that led us to distribute bids across multiple hours, whereas not all hours have the same profitability as hour 19.
    \item Strategy \textbf{T7} has a significant margin over other strategies, indicating the high potential of trading opportunities at hour 19 and highlighting the importance of developing accurate price forecasts, especially for peak hours.
    \item All the strategies that are based on prediction results incur a significant dip on Aug. 20, 2024. It is attributed to a misprediction where the model incorrectly forecasted that the SCED price would be lower than the DAM price, leading to substantial losses. This highlights the risk associated with relying on model predictions for trading, particularly when the precision is not consistently high.
\end{itemize}

\begin{figure}
    \centering
    \includegraphics[width=1\linewidth]{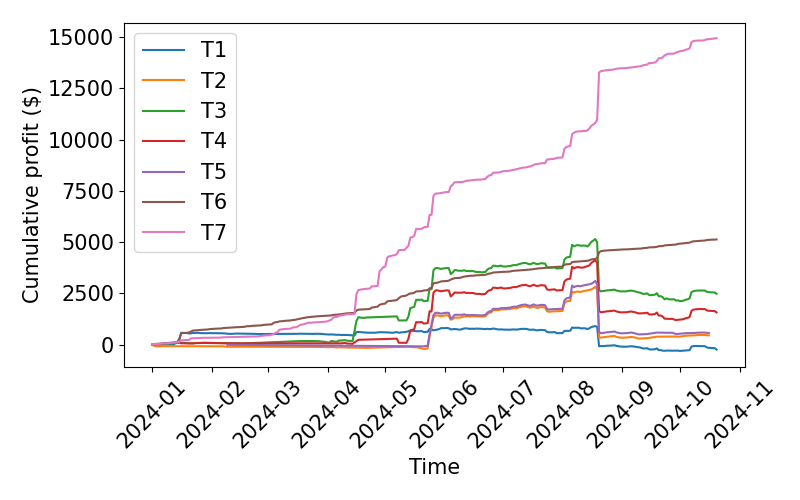}
    \caption{Cumulation profit for the best model using different trading strategies.}
    \label{cumulative_profit}
\end{figure}

\section{Conclusion and Future Work}

This paper presented a Transformer-based deep learning model for forecasting the price spread between the SCED and DAM prices in the ERCOT market, with a specific focus on enabling profitable virtual bidding strategies. The proposed model leverages advanced time-series features, such as load forecasts and renewable generation forecasts, and employs a walk-forward validation approach to adapt to changing market conditions. The results demonstrated that the model can provide valuable predictions, particularly for hour 19, where trading strategies based on the model's output at this hour yielded the highest cumulative profit. The proposed method highlighted the importance of evaluating the electricity price forecast model from the perspective of virtual bidding, while the metrics of profit and classification do not always align.

There are some limitations to the current work. Current input features, while informative, may not capture all the factors influencing price spreads, such as market sentiment, fuel prices, unexpected grid events, or weather data. The lack of such additional contextual information may lead to reduced model accuracy during volatile periods. Experimenting with other deep learning architectures, such as LSTM or graph neural networks, may further improve the model's ability to capture complex dependencies in the data. The significant losses observed during certain periods highlight the limitations of the current approach. Enhancing the model's robustness and incorporating risk management measures are critical areas for future work. Future improvements could include employing ensemble methods to mitigate the impact of mispredictions.

\bibliographystyle{IEEEtran}
\bibliography{references}

\end{document}